\documentclass[10pt,twocolumn,letterpaper]{article}

\usepackage[pagenumbers]{cvpr} % To force page numbers, e.g. for an arXiv version

\usepackage{amsmath}
\usepackage{graphicx}
\usepackage{wrapfig}
\usepackage{algorithm}
\usepackage{algorithmic}
\usepackage{multirow}
\usepackage{graphicx}
\usepackage{amssymb}

\makeatletter
\renewcommand{\paragraph}{\@startsection{paragraph}{4}{\z@}%
  {1.5ex \@plus .5ex \@minus .2ex}%
  {-1em}%
  {\normalfont\normalsize\bfseries}}
\makeatother

\definecolor{cvprblue}{rgb}{0.21,0.49,0.74}
\usepackage[pagebackref,breaklinks,colorlinks,allcolors=cvprblue]{hyperref}

\title{LiveSync: Iterative Self-Training for Lip-Sync Generation}

\author{
Huaishao Luo\quad Huan Zhang$^{*}$\quad Guanbo Bao\quad Yinglu Liu \\
Youzheng Wu\quad Xiaodong He\\\\
JD Technology
}

\begin{document}
\maketitle
\iftoggle{cvprfinal}{\begingroup\renewcommand{\thefootnote}{*}\begin{NoHyper}\footnotetext{Project Leader}\end{NoHyper}\endgroup}{}
\begin{abstract}
Audio-driven lip synchronization aims to match target speech through localized mouth and jaw articulation while preserving subject identity, head pose, and surrounding facial context. Naturally recorded videos provide synchronized reconstruction targets but rarely offer pairs that are matched in identity and pose yet differ primarily in speech-related articulation. Mask-based training enables reconstruction from these videos, but random-reference inputs differ from the aligned self-reference used at inference.
We present \textit{LiveSync}, an iterative self-training framework in which a single mask-based model generates pseudo pairs and is refined for deployment. The model is first trained on a curated real-video corpus augmented with externally generated clips covering under-represented visual conditions. A frozen checkpoint then uses independently sampled audio to generate mouth-edited references from real-video frames, copying pixels outside the edit mask from each source frame. After confidence filtering, these references are paired with aligned masked inputs and original real-frame targets. The retained pseudo pairs are mixed with real training data to refine the same model from its current weights, updating training supervision without adding inference stages. Although we instantiate the framework with a single-pass GAN-based model, the self-training strategy could in principle also be adapted to diffusion-based models.
Experimental results show that LiveSync achieves leading synchronization performance on HDTF and competitive performance compared with the baselines on the AIGC-LipSync subset, while maintaining low inference latency.
\end{abstract}
\section{Introduction}
\label{sec:intro}
\begin{figure}[t]
    \centering
    \begin{subfigure}{0.85\columnwidth}
        \centering
        \includegraphics[width=\linewidth]{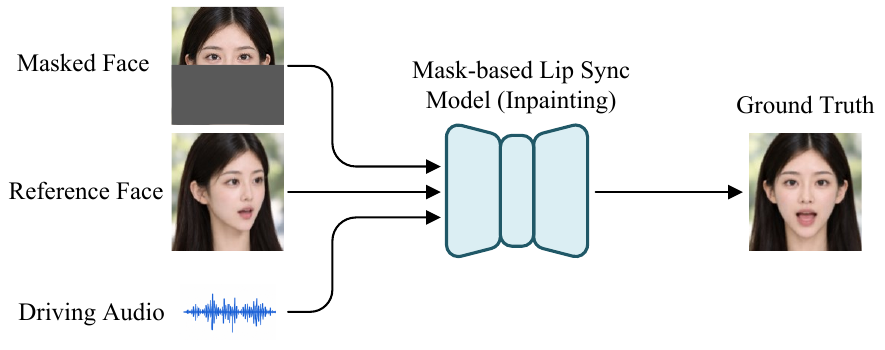}
        \caption{Mask-based Framework}
    \end{subfigure}

    \begin{subfigure}{0.95\columnwidth}
        \centering
        \includegraphics[width=\linewidth]{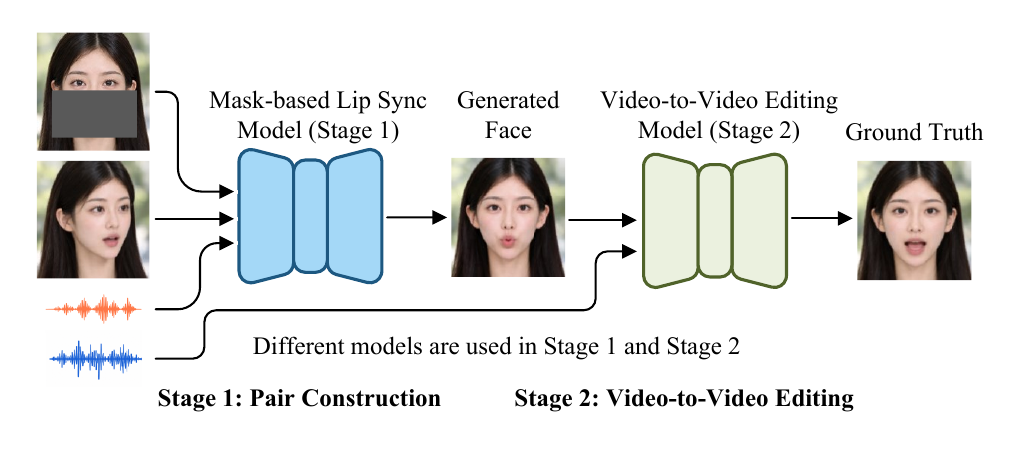}
        \caption{Two-stage Framework}
    \end{subfigure}

    \begin{subfigure}{0.99\columnwidth}
        \centering
        \includegraphics[width=\linewidth]{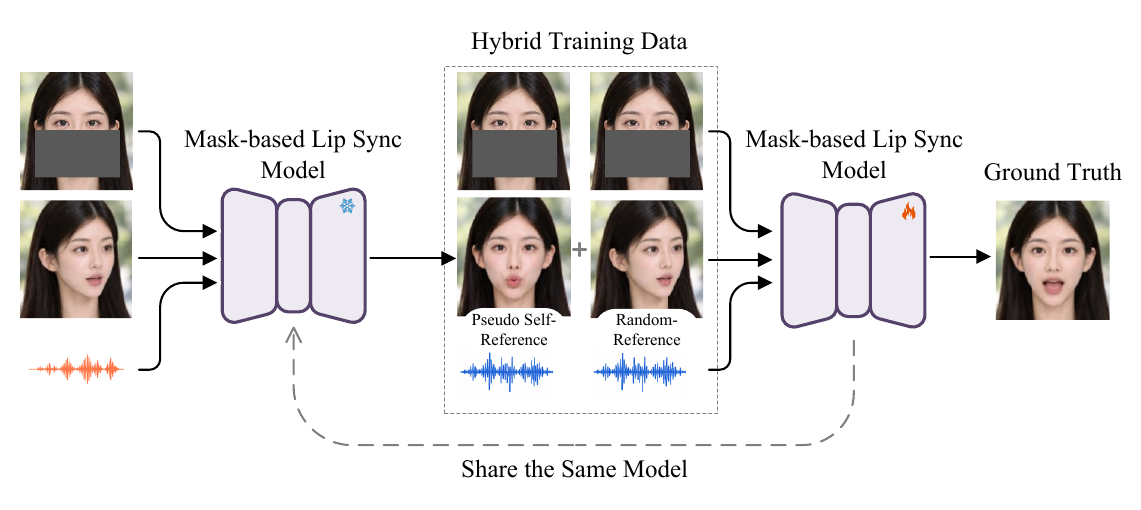}
        \caption{LiveSync: Iterative Self-training Framework}
    \end{subfigure}
    \caption{Comparison of lip-sync training paradigms. (a) Single-stage mask-based methods reconstruct masked facial regions from audio and reference frames. (b) Two-stage methods use a mask-based generator to construct pseudo pairs for a separate mask-free model. (c) LiveSync uses the same mask-based model to generate spatially aligned pseudo pairs, filter and mix them with real training data, and continue training from the previous round's weights, forming a closed-loop self-training pipeline. Gray-filled rectangles indicate masked regions.}
    \label{fig:introduction}
\end{figure}
Lip synchronization edits speech-related mouth and jaw motion in a source video to match target speech while retaining identity, head pose, surrounding facial context, and background~\cite{prajwal_lip_2020}. It supports video dubbing, virtual avatars, and audio-driven video editing, but remains challenging in the wild, particularly under large pose variation and occlusion.

Earlier audio-driven facial animation and lip-sync systems often separate motion prediction from image synthesis: audio is mapped to explicit facial geometry or parameters, or to learned motion representations, which then drive rendering or compositing~\cite{suwajanakorn_synthesizing_2017,lahiri_lipsync3d_2021,zhang_sadtalker_2023}. End-to-end models instead synthesize the edited facial region directly from audio and visual inputs. Ideally, these models would be trained on source--target pairs that differ only in speech-related mouth motion while preserving the source's other visual attributes. Naturally recorded videos provide synchronized audio and frames, but rarely contain such controlled editing pairs~\cite{zhang_syncanyone_2026,he_inpainting_2026}.

Mask-based methods obtain a reconstruction surrogate for this missing supervision. They mask the mouth-centered region of a real frame and reconstruct it from the corresponding audio and source or reference frames, using the original frame as the target~\cite{prajwal_lip_2020,zhang_dinet_2023,cheng_videoretalking_2022,zhang_musetalk_2025,bigata_keysync_2025}. This makes ordinary synchronized videos usable for training, but removes local context rather than providing intact source--target editing pairs. Moreover, to avoid exposing the target mouth during training, the masked target is commonly paired with a randomly sampled same-identity reference. That reference may differ in pose and appearance, whereas inference uses the source frame and its spatially aligned masked self. The resulting context loss and training--inference mismatch complicate precise mouth editing, especially under extreme poses and occlusion~\cite{zhang_syncanyone_2026,he_inpainting_2026}.

Recent methods address the supervision gap through mask-free or hybrid editing. OmniSync~\cite{peng_omnisync_2025} explores mask-free editing with Diffusion Transformers, while UniSync~\cite{fan_unisync_2026} combines mask-free pose-anchored training with mask-based blending at inference. SyncAnyone~\cite{zhang_syncanyone_2026} and X-Dub~\cite{he_inpainting_2026} explicitly construct the missing supervision: a mask-based generator synthesizes pseudo pairs, which train a separate mask-free model. These methods shift the emphasis from reconstructing missing pixels to changing mouth articulation while preserving the source performance.

Pretrained generative priors and paired editing supervision play complementary roles. Large-scale diffusion pretraining provides appearance and geometry priors that can support synthesis across diverse poses and partial occlusions, but visual plausibility alone does not ensure accurate audio-driven editing. Pseudo pairs that retain aligned source context while varying speech-related articulation provide a targeted signal for learning the audio-to-lip mapping. In mask-free models, complete source context additionally constrains where and how the image should change~\cite{peng_omnisync_2025,zhang_syncanyone_2026,he_inpainting_2026}. These constraints are especially valuable under large pose variation and occlusion, where insufficiently constrained generation may introduce unsupported details or alter content that should remain unchanged. The two-stage approaches above, however, use the mask-based model as a data generator rather than continuously refining it as the final model.

We ask whether better editing supervision can instead improve the same mask-based model. We introduce \textit{LiveSync}, a data-centric iterative self-training framework in which the same mask-based model generates pseudo pairs and is refined across training rounds. The base model is trained on a curated real-video corpus augmented with externally generated clips for under-represented visual conditions. A frozen checkpoint is then driven by randomly sampled audio to synthesize references with altered mouth motion while preserving the original real frame outside the edit mask. Pairing these references with the original real frames yields spatially aligned pseudo pairs with real reconstruction targets. We filter generated references for confidence, mix the retained pairs with real training data, and continue optimizing the same model from its previous-round weights, forming the closed loop in \cref{fig:introduction}.

Unlike random references, these self-generated references share the target frame's pose and surrounding appearance, reducing unrelated visual differences and matching the self-reference configuration used at inference. The aligned supervision is designed to strengthen audio-driven mouth control while preserving source-controlled content, making it particularly relevant to the pose and occlusion challenges above. Filtering limits the influence of unreliable generated references, while real-data mixing retains reference diversity. All refinement directly updates the deployment model, without a hand-off to a separate mask-free model.

We instantiate LiveSync with a single-pass GAN model. Pseudo-pair synthesis and filtering occur only during training and add no inference stages. Our focus is on supervision construction and reuse rather than a particular generator family. Applying the same principle to diffusion-based models remains future work.

Our contributions are threefold. First, we construct spatially aligned pseudo pairs with real reconstruction targets to reduce the mismatch between random-reference training and self-reference inference. Second, we develop a closed-loop training process that combines confidence filtering and mixed-data training to refine the same model across rounds. Third, we instantiate the framework using a single-pass GAN-based model and validate the effectiveness of self-training via benchmark comparisons and ablation studies on real and AI-generated videos. Despite operating at three times the linear face resolution of the fastest method, LiveSync maintains low inference latency and ranks as the second-fastest among the compared methods.

\section{Related Work}
\label{sec:related_work}

\subsection{Mask-based Lip Sync}
Mask-based lip-sync methods typically reconstruct a masked mouth region or edit local facial content conditioned on audio and visual references. Early systems such as Wav2Lip~\cite{prajwal_lip_2020} and VideoReTalking~\cite{cheng_videoretalking_2022} establish effective pipelines for in-the-wild videos, while DINet~\cite{zhang_dinet_2023} combines deformation and inpainting to preserve fine facial details in high-resolution dubbing. Later work targets fidelity, efficiency, or robustness. MuseTalk~\cite{zhang_musetalk_2025} uses latent-space inpainting and spatio-temporal sampling for real-time generation, whereas KeySync~\cite{bigata_keysync_2025} addresses expression leakage and occlusion through masking. Diffusion-based methods, including Diff2Lip~\cite{mukhopadhyay_diff2lip_2023}, SayAnything~\cite{ma_sayanything_2025}, LatentSync~\cite{li_latentsync_2025}, and UniSync~\cite{fan_unisync_2026}, explore improved realism and broader operating conditions. Lip Forcing~\cite{cho_lip_2026} distills a bidirectional video diffusion teacher into few-step autoregressive students, while EfficientSync~\cite{hong_efficientsync_2026} focuses on efficient reference-texture transfer through deformation-based feature fusion and adaptive masking.

Across these methods, progress has primarily come from changes to the generator, reference use, or synchronization objective. LiveSync instead focuses on the supervision available to the model. It constructs spatially aligned pseudo pairs and uses them to iteratively refine the same mask-based model, reducing the gap between randomly referenced training examples and the aligned self-reference input used at inference.

\subsection{Pseudo-paired Supervision for Lip Sync}
Naturally recorded videos rarely provide pairs that share closely matched identity and pose but differ in speech-related articulation, motivating pseudo-paired supervision for lip-sync editing. SyncAnyone~\cite{zhang_syncanyone_2026} and X-Dub~\cite{he_inpainting_2026} address this gap with two-stage pipelines in which a mask-based model first generates audio-driven mouth motion, and its outputs then provide pseudo pairs for training a separate mask-free model. TBDub~\cite{li_tbdub_2026} extends this strategy to production-domain footage, with additional attention to domain adaptation and diffusion acceleration. OmniSync~\cite{peng_omnisync_2025} also adopts mask-free editing, using a diffusion transformer for direct frame editing across diverse visual conditions, but follows a distinct training paradigm.

Together, these works illustrate distinct routes to full-context editing, including pseudo-pair bootstrapping and direct mask-free diffusion-based editing. LiveSync shares the data-centric motivation of pseudo-paired methods, but uses filtered generated references together with real target frames to iteratively refine the mask-based model itself. Unlike two-stage bootstrapping methods that hand off supervision to a separately trained mask-free model, LiveSync keeps supervision construction and model refinement within a single training loop.

\section{Method}
\label{sec:method}

\begin{figure*}[t]
    \centering
    \includegraphics[width=0.90\linewidth, keepaspectratio]{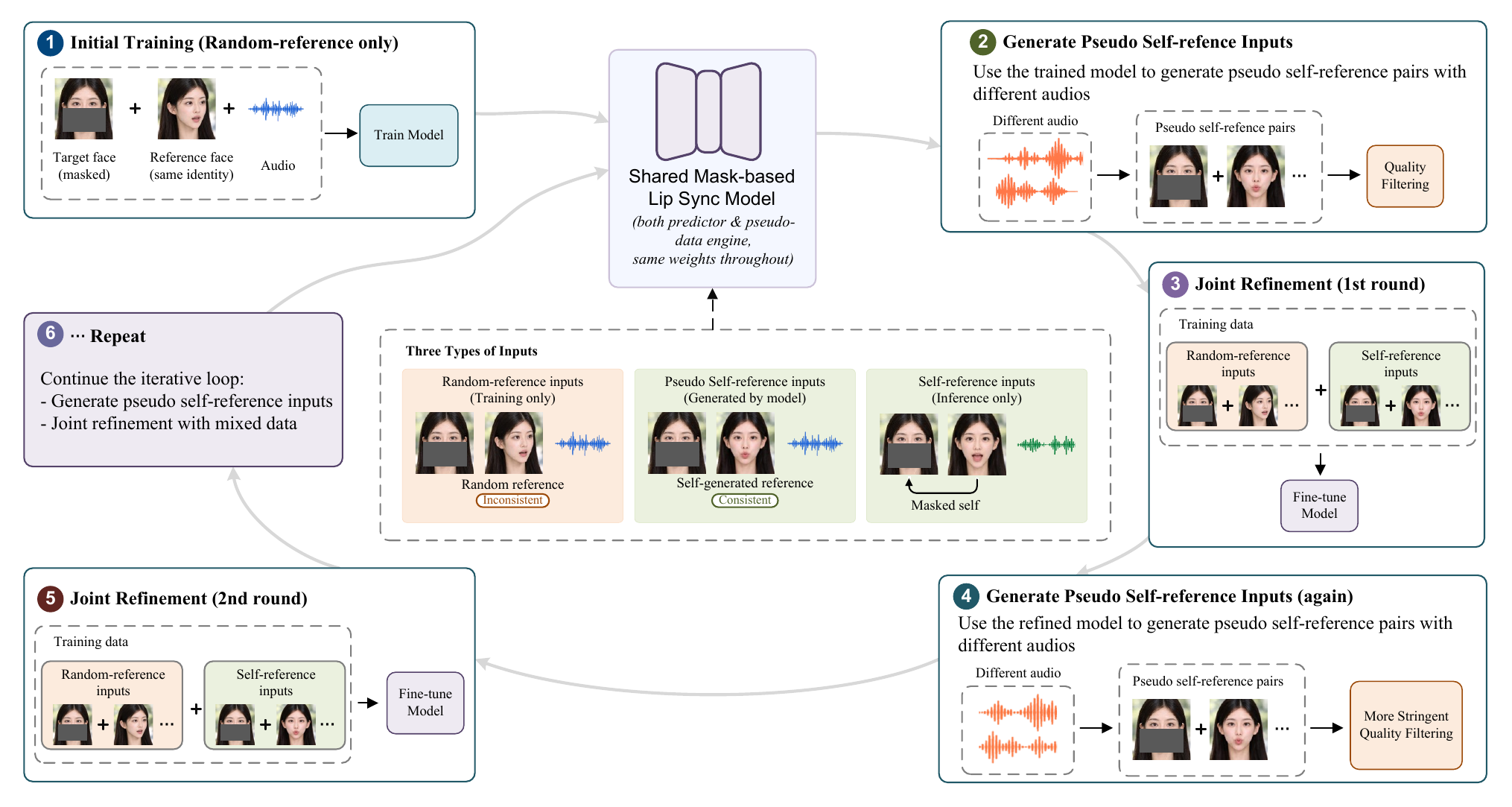}
    \caption{Overview of LiveSync's closed-loop self-training. The figure highlights the mismatch between inference, which uses a frame and its aligned masked self, and conventional training, which uses a randomly sampled reference. A frozen checkpoint driven by sampled audio produces a self-generated reference that preserves the target frame outside the edit mask. Confidence-filtered aligned pairs are then mixed with real pairs to refine the same model.}
    \label{fig:framework}
\end{figure*}

\subsection{Problem Setup}
Given a talking-face video with $T$ frames, $X=\{x_t\}_{t=1}^{T}$, and target speech $A$, lip synchronization generates $\hat{Y}=\{\hat{y}_t\}_{t=1}^{T}$ whose speech-related mouth and jaw motion matches $A$ while maintaining identity and head pose, limiting unintended changes to surrounding appearance, and preserving temporal coherence. Following common practice~\cite{prajwal_lip_2020,cheng_videoretalking_2022}, we detect and crop the face in each frame. For each crop of height $H$ and width $W$, we define a face-relative binary mask $m_t\in\{0,1\}^{H\times W}$ in crop coordinates. Here, $m_t=1$ marks the mouth region to be edited and $m_t=0$ marks pixels to preserve. The mask is broadcast across color channels. Ideally, training pairs would share identity and pose while differing primarily in speech-related articulation, but naturally recorded videos rarely provide such pairs~\cite{zhang_syncanyone_2026,he_inpainting_2026}.

At inference, a frame and its masked self are spatially aligned. During conventional mask-based training, the target mouth is masked in the input, and the masked target is paired with a randomly sampled reference of the same identity. Using the unmasked target itself as the reference would expose its ground-truth mouth shape, allowing the model to copy the target articulation rather than infer it from audio. A random reference avoids this leakage but may differ in pose, expression, or illumination, creating a training--inference input mismatch.

LiveSync addresses this mismatch by using the trained model as a \emph{self-improving data engine}. A frozen checkpoint driven by sampled audio synthesizes a self-generated reference. Pairing it with the original masked frame yields an aligned pseudo pair with a real target. Confidence filtering and iterative training on real and pseudo pairs are detailed below. Unlike two-model bootstrapping, the same model is refined throughout, without a separate mask-free student.

\subsection{Base Mask-Based Lip-Sync Model}
LiveSync uses a deterministic, single-pass GAN-based model with a pretrained VAE encoder $E_\phi$ and decoder $V_\omega$ for latent encoding and reconstruction. The audio-conditioned latent predictor $F_\eta$ takes a reference frame, a masked target frame, and an audio window and predicts an edited visual latent. The VAE decoder reconstructs the face crop, with reference features injected at intermediate decoder blocks as described in the training objective. We use $G_\theta$ for this complete editing model, with parameters $\theta$ covering the predictor, decoder, and feature projections. Unlike a diffusion pipeline, it predicts and decodes in one pass without iterative denoising. Two region-specific critics are used during training, as described in \cref{sec:training_objective}.

At inference, for source frame $x_t$, the masked self is
\begin{equation}
\tilde{x}_t = (1-m_t) \odot x_t,
\end{equation}
where $\odot$ denotes element-wise multiplication. Given the target audio window $A'_t$, the model predicts the edited crop $G_\theta(x_t,\tilde{x}_t,A'_t)$. We paste only the mouth region from this prediction into the source frame.
\begin{equation}
\hat{y}^{\mathrm{inf}}_t = (1-m_t) \odot x_t + m_t \odot G_\theta(x_t, \tilde{x}_t, A'_t).
\end{equation}
Thus, pixels outside the mask are copied unchanged from $x_t$.

For real-data training, let $y_t$ be a ground-truth frame with synchronized audio window $A_t$, and let $r_t$ be a randomly sampled reference frame of the same identity. Its masked target is $\tilde{y}_t=(1-m_t)\odot y_t$, and the composed prediction is
\begin{equation}
\hat{y}^{\mathrm{real}}_t = (1-m_t) \odot y_t + m_t \odot G_\theta(r_t, \tilde{y}_t, A_t).
\end{equation}
Here, $\hat{y}^{\mathrm{inf}}_t$ and $\hat{y}^{\mathrm{real}}_t$ denote the inference and real-training outputs, respectively. The former uses the aligned source frame as reference, whereas the latter uses a random reference to avoid exposing the target mouth. In the loss definitions below, $\hat{y}_t$ denotes the composed prediction for the current real or pseudo training example.

\subsection{Mask Construction and Boundary Handling}
\label{sec:mask_strategy}
\begin{figure*}[t]
    \centering
    \includegraphics[width=0.75\linewidth]{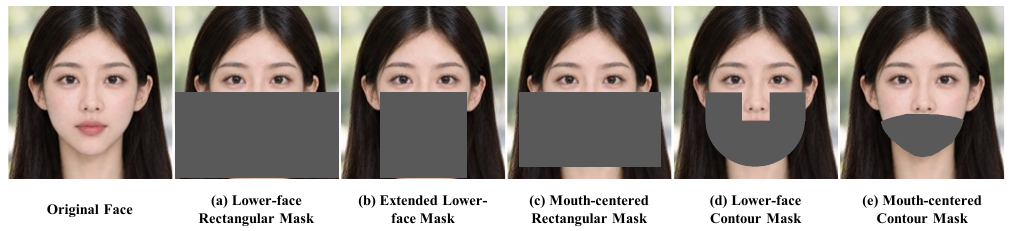}
    \caption{Comparison of representative face masking strategies. LiveSync adopts (c), the mouth-centered rectangular mask, which removes the target lip motion while retaining surrounding facial context. We further use boundary-aware padding when the face crop partially extends beyond the video frame.}
    \label{fig:mask_comparison}
\end{figure*}

Our mask is defined relative to the detected face crop rather than at a fixed position in the full video frame, so the crop and mask move together with the subject. As illustrated in \cref{fig:mask_comparison}, masks can cover the entire lower face or focus on the mouth, and can use either rectangular or contour-based boundaries. LiveSync adopts the mouth-centered rectangular mask in \cref{fig:mask_comparison}(c). Centering the mask on the detected mouth reliably removes the target lip motion, while the rectangular region provides sufficient surrounding facial context for reconstruction and efficient batched processing.

Faces near an image boundary require special treatment because the desired face box may extend beyond the valid frame. Directly clipping or resizing such a box changes its scale and spatial layout across frames, causing temporal jitter and geometric distortion. We instead preserve the original box size and fill all out-of-frame pixels with black. The padded crop is passed through the same model without changing its spatial resolution. After inference, we remove the padded margins and paste only the valid generated region back into the original frame. This boundary-aware procedure keeps the face and mask coordinates consistent over time, and is therefore particularly friendly to large subject motion and partial out-of-frame cases.

\subsection{Diversity-Oriented Synthetic Data Construction}
Self-training improves input alignment but cannot introduce visual conditions absent from the original corpus. To broaden the base training set, we augment it with externally generated clips for under-represented scenarios, including extreme head poses, partial occlusions, stylized avatars, and non-human characters. These clips are produced by image- and audio-conditioned talking-avatar models such as JoyStreamer~\cite{wang_joystreamer_2026}. We retain only clips with stable face tracking, adequate audio--visual synchronization, preserved identity or character appearance, and acceptable visual quality. By contrast, pseudo-pair construction uses real-video targets and therefore retains real reconstruction supervision. The two components play complementary but distinct roles. External generation broadens visual coverage, while the self-improving data engine provides aligned editing supervision without changing the single-pass inference path.

\subsection{Training Objective}
\label{sec:training_objective}
For each synchronized training tuple, we optimize the model with pixel, perceptual, lip-synchronization, and region-specific Wasserstein adversarial losses. The coefficients $\lambda_{\mathrm{rec}}$, $\lambda_{\mathrm{per}}$, $\lambda_{\mathrm{sync}}$, and $\lambda_{\mathrm{wgan}}$ weight the corresponding losses, while $\lambda_m$ weights the additional mouth-region reconstruction term. The reconstruction terms are
\begin{align}
\mathcal{L}_{\text{rec}} &= \|\hat{y}_t - y_t\|_1 + \lambda_{m}\|m_t \odot (\hat{y}_t-y_t)\|_1, \\
\mathcal{L}_{\text{per}} &= \text{LPIPS}(\hat{y}_t, y_t).
\end{align}

\paragraph{Lip-only SyncNet supervision.}
LatentSync demonstrates the importance of a well-converged SyncNet for explicitly learning audio--visual correspondence~\cite{li_latentsync_2025}. We use pretrained visual and audio encoders $s_v$ and $s_a$ as a fixed SyncNet supervisory network~\cite{chung2017out,prajwal_lip_2020}. SyncNet parameters remain unchanged, and its audio--visual similarity loss guides the generator through the predicted lip crops. Let $b$ index samples in a minibatch and $L$ denote the temporal extent of its input. For each target frame $y_{b,t}$, facial landmarks define a tight mouth box with width $w_{b,t}$, height $h_{b,t}$, and center $c_{b,t}$. We use these landmarks only to locate the corresponding region in the generated frame $\hat{y}_{b,t}$. A shared square crop size for the batch is
\begin{equation}
d_{\mathcal{B}} = \max_{b,t}\max(w_{b,t},h_{b,t}),
\end{equation}
which allows all mouth regions to be extracted in a single batched operation. The generated crop is resized to the fixed SyncNet input resolution,
\begin{equation}
\hat{p}_{b,t}=\operatorname{Resize}\!\left(\operatorname{Crop}(\hat{y}_{b,t},c_{b,t},d_{\mathcal{B}})\right).
\end{equation}
Stacking consecutive generated crops gives the lip clip $\hat{P}_{b,t:t+L}$. The synchronization loss compares its visual embedding with that of the corresponding audio segment,
\begin{equation}
\mathcal{L}_{\text{sync}} = 1-\cos\big(s_v(\hat{P}_{b,t:t+L}),s_a(A_{b,t:t+L})\big).
\end{equation}
Thus, $y_{b,t}$ supplies only the mouth-crop coordinates, while the loss is computed from the generated lip clip and its audio. Using a shared batch crop size preserves local lip detail and enables efficient tensor processing while limiting pose- and background-related shortcuts.

\paragraph{Encoder-initialized dual-region WGAN.}
We apply Wasserstein adversarial learning~\cite{arjovsky_wasserstein_2017,gulrajani_improved_2017} at two complementary spatial scales. The face critic $D_{\mathrm{face}}$ evaluates global structure, skin continuity, and blending over the lower-face region. The lip critic $D_{\mathrm{lip}}$ focuses on high-frequency lip contours, teeth, and inner-mouth texture within a mouth-centered crop. For an image $u$, let $R_r(u)$ extract region $r\in\{\mathrm{face},\mathrm{lip}\}$. Each critic takes this image crop as input and is initialized from the pretrained VAE encoder $E_{\phi}$. We retain the encoder's convolutional feature extractor and attach a scalar Wasserstein scoring head. The face and lip critics are initialized as separate copies and optimized independently. This initialization provides learned visual features from the start, avoiding a critic cold start while keeping adversarial feedback focused on the corresponding facial region.

Let $\bar{u}_r$ interpolate between real and generated crops, $R_r(y)$ and $R_r(\hat{y})$. We minimize the WGAN-GP critic objective
\begin{align}
\mathcal{L}_{D}^{r} ={}& \mathbb{E}\big[D_r(R_r(\hat{y}))\big]-\mathbb{E}\big[D_r(R_r(y))\big] \\
&+\lambda_{\mathrm{gp}}\mathbb{E}\left[\left(\|\nabla_{\bar{u}_r}D_r(\bar{u}_r)\|_2-1\right)^2\right],
\end{align}
while the generator receives
\begin{equation}
\mathcal{L}_{\text{wgan}}=-\alpha_{\mathrm{face}}\mathbb{E}[D_{\mathrm{face}}(R_{\mathrm{face}}(\hat{y}))]-\alpha_{\mathrm{lip}}\mathbb{E}[D_{\mathrm{lip}}(R_{\mathrm{lip}}(\hat{y}))].
\end{equation}
Here, $\lambda_{\mathrm{gp}}$ weights the gradient penalty, and $\alpha_{\mathrm{face}}$ and $\alpha_{\mathrm{lip}}$ balance the two critic terms. The face-level term discourages seams and inconsistent lower-face geometry, while the lip-level term promotes realistic speaking-region detail.

The complete editing objective is
\begin{equation}
\mathcal{L}_{\text{edit}} = \lambda_{\text{rec}}\mathcal{L}_{\text{rec}} +
\lambda_{\text{per}}\mathcal{L}_{\text{per}} +
\lambda_{\text{sync}}\mathcal{L}_{\text{sync}} +
\lambda_{\text{wgan}}\mathcal{L}_{\text{wgan}}.
\end{equation}

\paragraph{Late-stage VAE decoder refinement.}
Latent decoding can lose fine details, particularly around teeth, lip boundaries, and skin texture. To restore frame-specific appearance, we extract a multi-scale feature pyramid $\{f_t^{\ell}\}$ from the reference frame and inject it into intermediate VAE decoder blocks. At decoder level $\ell$, the hidden activation becomes
\begin{equation}
h_t^{\ell}=B_{\ell}(h_t^{\ell-1})+\gamma_{\ell}P_{\ell}(f_t^{\ell}),
\end{equation}
where $B_{\ell}$ is the decoder block, $P_{\ell}$ projects the reference feature to the decoder channel dimension, and $\gamma_{\ell}$ is a learnable modulation weight. These features restore cues omitted by latent compression and anchor the output to the reference appearance.

During initial training, we freeze $V_{\omega}$ and the reference-feature projections so that $F_{\eta}$ first learns stable audio-conditioned mouth motion. Once synchronization and reconstruction losses stabilize, we jointly fine-tune the predictor, decoder, and projections, using a smaller learning rate for the decoder. This delayed update limits premature drift of the pretrained VAE while improving local detail and mask-boundary transitions.

\subsection{Pseudo-Paired Data Synthesis}
After the base model $G_{\theta}^{(0)}$ converges, we construct aligned references for real-video targets. Given a real training tuple $(r_t,\tilde{y}_t,A_t,y_t)$, we sample an unrelated audio window $A'_t$ and use the frozen current checkpoint $G_{\theta}^{(r)}$ to synthesize a counterfactual reference
\begin{equation}
s_t = (1-m_t) \odot y_t + m_t \odot G_{\theta}^{(r)}(r_t, \tilde{y}_t, A'_t).
\end{equation}
The generated frame $s_t$ follows the randomly sampled audio in its mouth region, while pixels outside the edit mask are copied exactly from $y_t$. Consequently, masking $s_t$ gives $(1-m_t)\odot s_t=\tilde{y}_t$. We can therefore form the pseudo training tuple $(s_t,\tilde{y}_t,A_t,y_t)$. Here, $s_t$ is the self-generated unmasked reference, $\tilde{y}_t$ is its spatially aligned masked self, and the original audio $A_t$ and real frame $y_t$ provide the supervision target. Unlike the original tuple with random reference $r_t$, this construction matches the self-reference input pattern used at inference while retaining a real ground-truth target.

A naive self-training loop can amplify teacher errors, so we retain only high-confidence self-generated references. We assess confidence using lip-sync similarity $q_{\text{sync}}$ between $s_t$ and the sampled audio $A'_t$ with SyncNet~\cite{chung2017out,prajwal_lip_2020}, identity similarity $q_{\text{id}}$ between $s_t$ and $y_t$ with ArcFace~\cite{deng2019arcface}, and blind image quality $q_{\text{iqa}}$ of $s_t$ with HyperIQA~\cite{su2020blindly}. The thresholds $\tau_{\text{sync}}$, $\tau_{\text{id}}$, and $\tau_{\text{iqa}}$ are preset minimum scores. A sample is retained only if $q_{\text{sync}} > \tau_{\text{sync}}$, $q_{\text{id}} > \tau_{\text{id}}$, and $q_{\text{iqa}} > \tau_{\text{iqa}}$.
We also discard trivial examples with negligible mouth change, which avoids over-populating the pseudo corpus with near-identity copies.

\subsection{Iterative Self-Training}
Let $\mathcal{P}_{\mathrm{real}}$ and $\mathcal{P}_{\mathrm{pseudo}}^{(r)}$ be the empirical sampling distributions over the original real tuples and the filtered pseudo tuples from round $r$, respectively. We train on their mixture.
\begin{equation}
\mathcal{P}_{\mathrm{mix}}^{(r)}=(1-\beta_r)\mathcal{P}_{\mathrm{real}}+\beta_r\mathcal{P}_{\mathrm{pseudo}}^{(r)},
\end{equation}
where $\beta_r\in[0,1]$ is the pseudo-tuple sampling proportion. Each round starts from the current model weights. A frozen copy generates pseudo pairs, and the trainable model is then optimized on the mixture to produce the next checkpoint. This retains weight continuity rather than handing supervision to a separately trained model.

Real tuples $(r_t,\tilde{y}_t,A_t,y_t)$ and pseudo tuples $(s_t,\tilde{y}_t,A_t,y_t)$ share the same real reconstruction target, so both use the same editing objective. We apply mild mask jitter, color augmentation, and audio-window resampling during training. Real tuples retain reference diversity, while pseudo tuples expose the model to the aligned self-reference input used at inference. The update changes the training data, not the single-pass inference path.
\section{Experiments}
\label{sec:experiments}

\subsection{Experimental Settings}
\paragraph{Datasets.}
We evaluate LiveSync on HDTF~\cite{zhang2021flow} and the AIGC-LipSync Benchmark introduced by OmniSync~\cite{peng_omnisync_2025}. HDTF contains real talking-face videos, while AIGC-LipSync covers AI-generated videos, including stylized and non-human subjects. To reduce evaluation time, we randomly select 100 AIGC-LipSync videos and use the same subset for all methods. All methods use the same selected videos and target audio in our evaluation runs. LiveSync is trained on a curated real-video corpus augmented with externally generated clips for under-represented visual conditions, as described in \cref{sec:method}.

\paragraph{Implementation details.}
LiveSync is a deterministic, single-pass GAN model that uses a pretrained VAE for latent encoding and decoding, with a U-Net-style audio-conditioned predictor and region-specific critics. Multi-scale reference-frame features are injected into intermediate decoder blocks during final refinement. We use an audio window of $K{=}5$ frames on each side.

\paragraph{Metrics.}
We report Sync-C and Sync-D for audio--visual synchronization~\cite{chung2017out,prajwal_lip_2020}, PSNR and SSIM for frame-level similarity~\cite{wang2004image}, FID and FVD for image- and video-level distributional similarity~\cite{heusel2017gans,unterthiner2018towards}, LPIPS for perceptual similarity~\cite{zhang2018unreasonable}, and CSIM for face-identity similarity~\cite{deng2019arcface}. Higher Sync-C, PSNR, SSIM, and CSIM and lower Sync-D, FID, FVD, and LPIPS are better.

\subsection{Quantitative Comparison}
\begin{table*}[t]
    \centering
    \small
    \setlength{\tabcolsep}{5pt}
    \resizebox{\linewidth}{!}{%
    \begin{tabular}{lccccccccc}
        \toprule
        Method & \multicolumn{2}{c}{Lip Sync} & \multicolumn{5}{c}{Visual Quality} & \multicolumn{1}{c}{Identity}  & \multicolumn{1}{c}{Efficiency} \\
        \cmidrule(lr){2-3} \cmidrule(lr){4-8} \cmidrule(lr){9-9} \cmidrule(lr){10-10}
         & Sync-C $\uparrow$ & Sync-D $\downarrow$ & PSNR $\uparrow$ & SSIM $\uparrow$ & FID $\downarrow$ & FVD $\downarrow$ & LPIPS $\downarrow$ & CSIM $\uparrow$  & Latency ($\times$) $\downarrow$ \\
        \midrule
        \multicolumn{10}{c}{HDTF} \\
        \midrule
        Wav2Lip~\cite{prajwal_lip_2020} & \underline{8.419} & 7.158 & 27.609 & 0.8860 & 31.727 & 318.165 & 0.1848 & 0.4753  & \textbf{1.0} \\
        VideoReTalking~\cite{cheng_videoretalking_2022} & 8.199 & \underline{6.930} & 25.944 & 0.8506 & 33.741 & 310.598 & 0.2011 & 0.4379  & 9.2 \\
        IP-LAP~\cite{zhong_iplap_2023} & 4.544 & 9.992 & 29.035 & 0.8952 & 20.623 & 243.397 & 0.1565 & 0.4936  & 143.9 \\
        DINet~\cite{zhang_dinet_2023} & 6.787 & 8.118 & 26.507 & 0.8819 & 14.447 & 151.289 & 0.1507 & 0.4797  & 18.7 \\
        MuseTalk~\cite{zhang_musetalk_2025} & 7.020 & 7.999 & 28.663 & 0.9086 & 11.400 & 151.682 & 0.1207 & 0.5468  & 18.5 \\
        LatentSync~\cite{li_latentsync_2025} & 6.810 & 7.776 & \textbf{31.223} & \textbf{0.9444} & \underline{8.038} & \textbf{109.404} & \textbf{0.0860} & \textbf{0.5987}  & 265.4 \\
        X-Dub~\cite{he_inpainting_2026} & 4.996 & 10.112 & 23.687 & 0.7663 & 11.450 & 158.650 & 0.2016 & 0.3501  & 987.9 \\
        Lip Forcing~\cite{cho_lip_2026} & 6.129 & 8.819 & \underline{29.817} & \underline{0.9203} & \textbf{7.321} & \underline{131.012} & \underline{0.1094} & 0.5680  & 143.5 \\
        TBDub~\cite{li_tbdub_2026} & 7.078 & 8.414 & 26.190 & 0.8715 & 14.961 & 142.123 & 0.1421 & 0.4965  & 48.3 \\
        LiveSync & \textbf{9.138} & \textbf{6.395} & 28.921 & 0.9144 & 16.589 & 142.584 & 0.1113 & \underline{0.5759}  & \underline{3.1} \\
        \midrule
        \multicolumn{10}{c}{AIGC-LipSync Benchmark} \\
        \midrule
        Wav2Lip~\cite{prajwal_lip_2020} & \underline{5.887} & 8.499 & 27.929 & 0.8848 & 44.829 & 401.002 & 0.2527 & 0.4496  & \textbf{1.0} \\
        VideoReTalking~\cite{cheng_videoretalking_2022} & 5.630 & 8.309 & 25.372 & 0.8639 & 54.387 & 320.042 & 0.2484 & 0.4290  & 9.2 \\
        IP-LAP~\cite{zhong_iplap_2023} & 3.305 & 10.044 & 28.369 & 0.9021 & 41.838 & 253.867 & 0.2073 & 0.4728  & 143.9 \\
        DINet~\cite{zhang_dinet_2023} & 3.059 & 10.393 & 22.666 & 0.8641 & 39.439 & 326.767 & 0.2324 & 0.4533  & 18.7 \\
        MuseTalk~\cite{zhang_musetalk_2025} & 3.687 & 9.800 & 26.756 & 0.9142 & 26.768 & 271.187 & 0.1636 & 0.5214  & 18.5 \\
        LatentSync~\cite{li_latentsync_2025} & \textbf{5.920} & \textbf{8.039} & \textbf{31.016} & \textbf{0.9462} & 20.494 & \textbf{135.993} & \textbf{0.1148} & \textbf{0.5582}  & 265.4 \\
        X-Dub~\cite{he_inpainting_2026} & 4.373 & 9.579 & 24.557 & 0.8317 & 18.193 & 240.506 & 0.2059 & 0.4142  & 987.9 \\
        Lip Forcing~\cite{cho_lip_2026} & 5.068 & 8.646 & 28.197 & 0.9131 & 22.818 & \underline{158.803} & 0.1428 & 0.5314  & 143.5 \\
        TBDub~\cite{li_tbdub_2026} & 3.944 & 9.922 & 29.422 & 0.9164 & \textbf{14.687} & 162.735 & \underline{0.1419} & 0.5160  & 48.3 \\
        LiveSync & 5.746 & \underline{8.251} & \underline{30.863} & \underline{0.9317} & \underline{15.370} & 181.594 & 0.1423 & \underline{0.5406}  & \underline{3.1} \\
        \bottomrule
    \end{tabular}}
    \caption{Quantitative comparison of LiveSync and all baselines on HDTF and a shared 100-video subset of AIGC-LipSync~\cite{peng_omnisync_2025}. All entries are measured under our evaluation protocol. PSNR is reported in dB. Latency is normalized to Wav2Lip (1.0$\times$) and repeated in both dataset blocks. Bold and underlining indicate the best and second-best values, respectively, for each metric within a dataset.}
    \label{tab:main_results}
\end{table*}

\Cref{tab:main_results} compares LiveSync with all baselines on HDTF and the selected 100-video AIGC-LipSync subset.

\paragraph{Results on HDTF.}
LiveSync achieves the best lip-synchronization scores, with Sync-C of 9.138 and Sync-D of 6.395. The strongest competing scores are 8.419 in Sync-C from Wav2Lip and 6.930 in Sync-D from VideoReTalking. LiveSync also outperforms both methods on all five visual-quality metrics and achieves the second-highest CSIM (0.5759). Compared with LatentSync, however, LiveSync has stronger synchronization scores, while LatentSync performs better on all five visual-quality metrics and CSIM. Lip Forcing achieves the lowest FID. These results identify synchronization as LiveSync's main strength on HDTF, while also showing a gap to the leading methods in visual quality.

\paragraph{Results on AIGC-LipSync.}
On the selected AIGC-LipSync subset, LiveSync ranks second in Sync-D and third in Sync-C, while also ranking second in PSNR, SSIM, FID, and CSIM. Its PSNR is 0.153\,dB lower than LatentSync's (30.863 vs.\ 31.016), whereas its FID is lower (15.370 vs.\ 20.494). TBDub is the only method with a lower FID (14.687), but LiveSync outperforms it on both synchronization metrics. LatentSync retains an advantage in FVD (135.993 vs.\ 181.594) and LPIPS (0.1148 vs.\ 0.1423). Thus, LiveSync is competitive in frame-level reconstruction and image-distribution similarity on this subset, but trails LatentSync in video-level distributional and perceptual similarity.

\paragraph{Inference latency.} We compare model-side latency under a deployment-oriented protocol, excluding components that can be processed offline (e.g., the VAE encoder and VideoReTalking's DNet and ENet) and timing only the model components required to generate the edited face. For diffusion baselines with multiple official configurations, we use the fastest official variant at its minimum official step when available, or the default step otherwise, and retain each method's default resolution. The evaluated face widths and heights range from 96 to 512 pixels, so latency is measured at each method's native resolution rather than at a common input size. As shown in \cref{tab:main_results}, latency is normalized to Wav2Lip (1.0$\times$). LiveSync takes 3.1$\times$ the Wav2Lip latency and is the second-fastest method, while operating at three times Wav2Lip's linear face resolution. This indicates a favorable speed--resolution trade-off under the reported protocol.

\subsection{Qualitative Comparison}
\begin{figure*}[t]
    \centering
    \includegraphics[width=\linewidth]{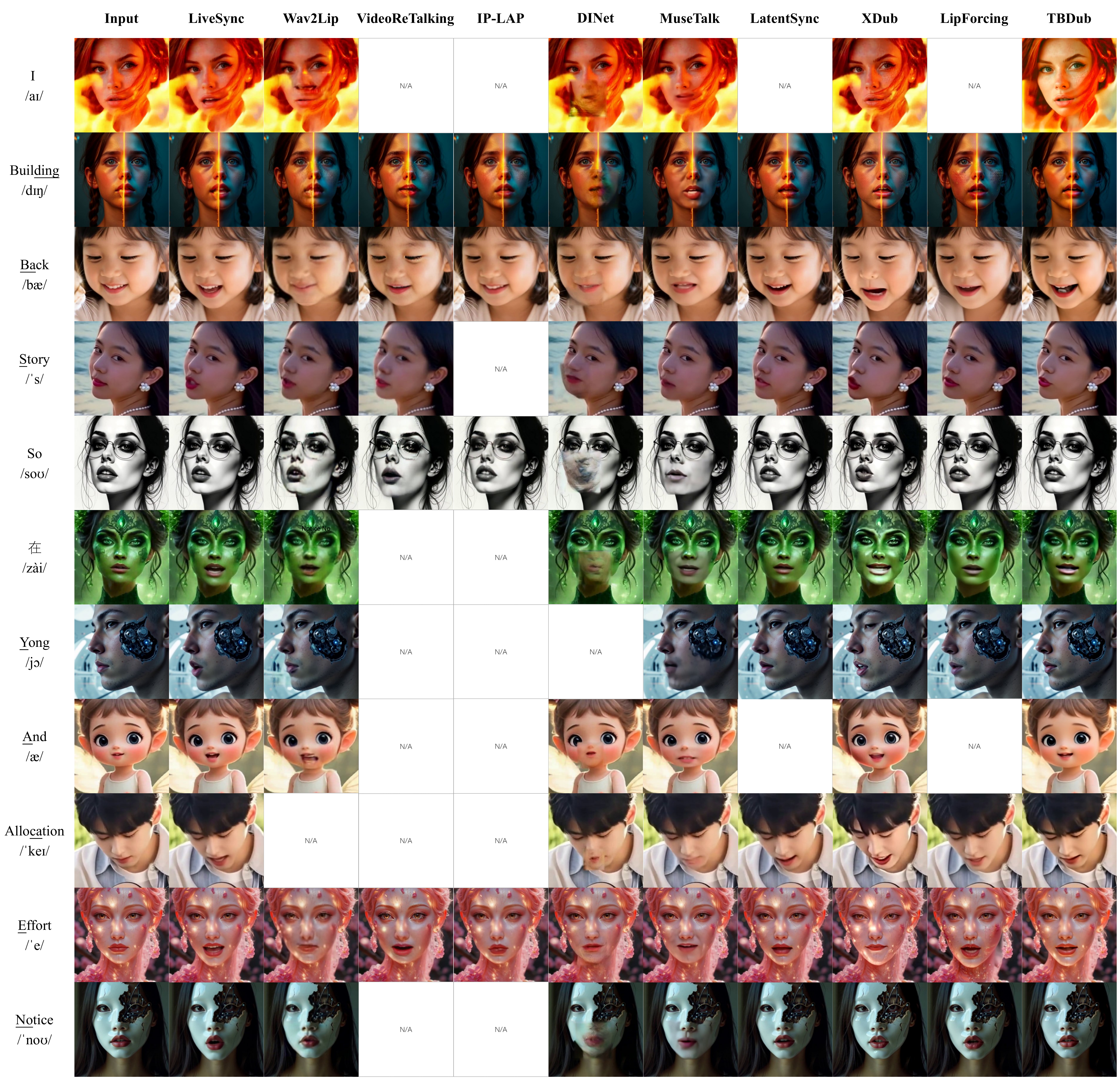}
    \caption{Qualitative comparison of lip-sync outputs. The first column shows the original input frame, followed by outputs from LiveSync and all baselines for the same target audio. Driving words and phonemes are listed at left, and ``N/A'' indicates that a method failed to generate an output. The examples include profile views, occluded faces, and stylized subjects.}
    \label{fig:qualitative_comparison}
\end{figure*}

Figure~\ref{fig:qualitative_comparison} compares selected source frames and outputs generated from the same target audio. The first column shows the original input. Subsequent columns show LiveSync and all baselines. Driving words and phonemes are listed at left, and ``N/A'' marks a generation failure. The examples include large pose changes, occlusions, and stylized faces, allowing qualitative inspection of articulation and preservation of source appearance.

Across the displayed driving phonemes, LiveSync, MuseTalk, X-Dub, and TBDub produce outputs for every case. Several other methods show ``N/A'' on one or more examples, including challenging pose and occlusion cases. The generated frames exhibit different mouth openings and lip configurations across target phonemes, providing qualitative evidence of articulation control. ``N/A'' denotes generation failure rather than a low-quality output.

The examples include side-facing subjects and partial occlusions, such as the cybernetic profile in ``Yong'' and the occluded face in ``Notice,'' as well as stylized characters. LiveSync produces outputs in these cases while retaining the visible pose, occluding elements, and character appearance outside the edited mouth region, suggesting robustness to the shown visual conditions.
\begin{table*}[t]
    \centering
    \small
    \setlength{\tabcolsep}{4pt}
    \resizebox{\linewidth}{!}{%
    \begin{tabular}{lcccccccc}
        \toprule
        Setting & Sync-C $\uparrow$ & Sync-D $\downarrow$ & PSNR $\uparrow$ & SSIM $\uparrow$ & FID $\downarrow$ & FVD $\downarrow$ & LPIPS $\downarrow$ & CSIM $\uparrow$ \\
        \midrule
        \multicolumn{9}{c}{HDTF} \\
        \midrule
        LiveSync & \textbf{9.138} & \textbf{6.395} & \textbf{28.921} & \textbf{0.9144} & \textbf{16.589} & \textbf{142.584} & \textbf{0.1113} & \textbf{0.5759} \\
        \quad w/o Pair & 8.938 & 6.566 & 27.461 & 0.8704 & 28.458 & 194.823 & 0.1656 & 0.4672 \\
        \quad w/o Pair \& GAN & 8.718 & 6.559 & 28.496 & 0.8924 & 32.175 & 251.520 & 0.1606 & 0.4998 \\
        \midrule
        \multicolumn{9}{c}{AIGC-LipSync Benchmark} \\
        \midrule
        LiveSync & \textbf{5.746} & 8.251 & \textbf{30.863} & \textbf{0.9317} & \textbf{15.370} & \textbf{181.594} & \textbf{0.1423} & \textbf{0.5406} \\
        \quad w/o Pair & 5.736 & \textbf{8.242} & 28.661 & 0.8923 & 24.219 & 234.588 & 0.1986 & 0.4686 \\
        \quad w/o Pair \& GAN & 5.153 & 8.556 & 29.825 & 0.9052 & 25.456 & 237.135 & 0.1929 & 0.4856 \\
        \bottomrule
    \end{tabular}}
    \caption{Cumulative ablation on HDTF and the shared 100-video AIGC-LipSync subset, using the eight metrics of the main comparison. The w/o Pair variant removes self-generated pseudo pairs, and w/o Pair \& GAN additionally removes adversarial training. All entries are measured results. Bold marks the best value among the three variants for each metric within each dataset.}
    \label{tab:ablation}
\end{table*}

Compared with the first column, LiveSync generally preserves source-controlled appearance while changing the mouth to match the target audio. This is particularly apparent in the third row (``Back''), where LiveSync closely preserves the child's framing and surrounding facial appearance while changing the articulation, appearing especially consistent with the input among the displayed outputs. These still frames support a qualitative comparison of appearance and mouth shape, but cannot establish temporal stability or audio--visual synchronization. Those properties require video inspection and quantitative metrics.

\subsection{Ablation Studies}
\begin{figure*}[t]
    \centering
    \includegraphics[width=\linewidth]{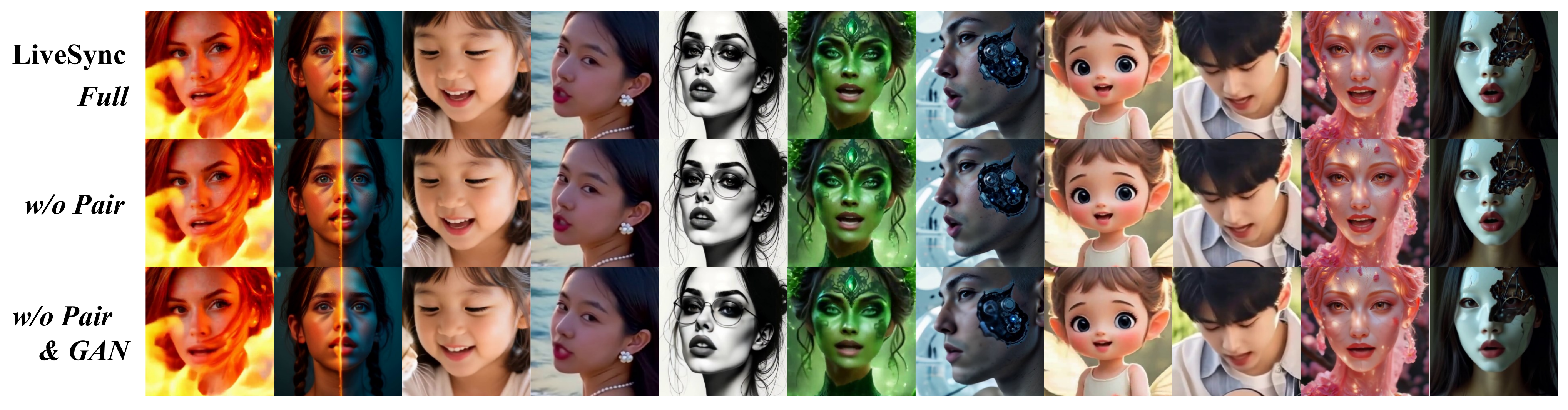}
    \caption{Qualitative comparison of the cumulative ablation variants. From top to bottom: full LiveSync, w/o Pair, and w/o Pair \& GAN, matching the configurations in \cref{tab:ablation}. Each column shows the same example across the three variants. The examples cover diverse facial appearances, stylized characters, and head poses, allowing comparison of mouth shape, lip boundaries, and inner-mouth appearance.}
    \label{fig:qualitative_ablation}
\end{figure*}

We conduct a cumulative ablation on HDTF and the same 100-video AIGC-LipSync subset used in the main comparison. We remove self-generated pseudo pairs while retaining GAN training (\emph{w/o Pair}), then remove adversarial training as well (\emph{w/o Pair \& GAN}), so the final variant disables both components. ``Pair'' denotes LiveSync's self-generated pseudo pairs, not real pairs or external clips. \Cref{tab:ablation} reports the scores, and \Cref{fig:qualitative_ablation} shows matched examples. The examples retain similar identity, pose, and overall appearance, with visible differences concentrated in mouth opening, lip contours, and inner-mouth detail, particularly for stylized or non-frontal subjects. Still images show appearance differences but cannot establish synchronization or temporal stability. The scores show clearer synchronization gains from pseudo-pair training on HDTF, while changes on AIGC-LipSync are small and mixed.

\paragraph{Effect of pseudo-paired supervision.}
Comparing LiveSync with \emph{w/o Pair} evaluates pseudo-paired supervision while retaining the GAN objective. On HDTF, the full model improves all eight metrics. For example, Sync-C rises from 8.938 to 9.138, while FID and FVD fall from 28.458 to 16.589 and from 194.823 to 142.584. PSNR, SSIM, LPIPS, and CSIM also improve. On AIGC-LipSync, seven metrics improve, with clear gains in PSNR (28.661 to 30.863), FID (24.219 to 15.370), and LPIPS (0.1986 to 0.1423). The synchronization changes there are small and mixed, with Sync-C changing from 5.736 to 5.746 and Sync-D from 8.242 to 8.251. Overall, the results are consistent with a broad visual-quality benefit from pseudo-paired supervision, whereas its synchronization benefit is not uniform across datasets.

\paragraph{Effect of adversarial training.}
Comparing \emph{w/o Pair} with \emph{w/o Pair \& GAN} assesses adversarial training only when pseudo pairs are absent. Retaining the GAN objective lowers FID and FVD on both datasets and improves Sync-C on both. It also improves Sync-D on AIGC-LipSync, while the HDTF difference in Sync-D is negligible and slightly favors the non-adversarial variant (6.559 vs. 6.566). Conversely, PSNR, SSIM, LPIPS, and CSIM favor the non-adversarial variant on both datasets. Thus, under this ablation, adversarial training trades pixel/perceptual and identity scores for better distributional metrics and, in most cases, synchronization scores. The full model gives the best value on all eight HDTF metrics and seven AIGC-LipSync metrics. The \emph{w/o Pair} variant has the best AIGC-LipSync Sync-D. The comparison between the second and third rows of \Cref{fig:qualitative_ablation} qualitatively illustrates how adversarial training affects mouth appearance when pseudo pairs are absent.

\section{Discussion: Avatar Generation versus Lip Sync}
\label{sec:discussion}

Image-and-audio-driven avatar generation and video lip synchronization both synthesize speech-conditioned mouth motion, but they impose different requirements on the learned data distribution. Avatar generation starts from a static or weakly constrained reference and synthesizes a complete motion trajectory. It is expected to coordinate head motion, facial expression, and sometimes body gestures with speech; rhythmic nods and emphasis-related movement are desirable, not spurious correlations~\cite{zhou_makelttalk_2020,zhang_sadtalker_2023,li_ditto_2025,wang_joystreamer_2026}. These models can also produce sharp, realistic, expressive outputs across diverse appearances, including stylized and non-human subjects. Lip-sync editing, in contrast, modifies speech-related mouth and jaw articulation within an existing performance while retaining its source-specific pose and surrounding motion.

Because target audio may be unrelated to the source performance, target articulation is ideally driven by target audio rather than inferred from source-correlated pose, expression, and body movement. Natural recordings rarely contain such counterfactual combinations: speech, expression, and gesture are statistically coupled. Models trained only on natural audio-video pairs can therefore exploit these correlations instead of learning a precise, independently controllable mapping from speech to lip motion~\cite{li_latentsync_2025,zhang_syncanyone_2026}.

The tasks also differ in who controls time. Avatar models generate and evolve their own motion; lip-sync models inherit a trajectory that may contain pose changes, frozen or duplicated frames, tracking failures, occlusions, and shot cuts. Temporal consistency therefore benefits from adapting to source continuity: strict framewise fidelity can cause flicker, while indiscriminate smoothing can alter appearance or propagate errors across cuts. Robust editing requires balancing fidelity and coherence according to the reliability of the input, not merely the generated history~\cite{bigata_keysync_2025,fan_unisync_2026}.

This distinction explains the role of factorized motion representations in earlier talking-face systems. Synthesizing Obama aligns predicted mouth shapes to the target video's 3D pose~\cite{suwajanakorn_synthesizing_2017}; LipSync3D separates geometry, pose, texture, and illumination~\cite{lahiri_lipsync3d_2021}; and SadTalker models expression separately from head pose~\cite{zhang_sadtalker_2023}. These explicit or learned factorizations reduce interference between audio-driven articulation and source-controlled motion. Modern pixel- and latent-space models may dispense with explicit 3D rendering, but not with the underlying identifiability problem.

Accordingly, LiveSync uses avatar-generated videos and pseudo-paired lip-sync data for complementary purposes. Large-scale generation broadens coverage of rare identities, extreme poses, occlusions, stylized characters, and non-human appearances, but largely retains natural coupling between speech and whole-body motion. Pseudo pairs instead combine a fixed source performance with independently sampled speech, using the unchanged context outside the edit mask to provide counterfactual supervision. Robust lip synchronization thus needs both broad visual coverage and deliberately decorrelated audio-motion supervision.
\section{Conclusion}
\label{sec:conclusion}

We introduced LiveSync, an iterative self-training framework in which a mask-based lip-sync model generates aligned pseudo-references for its own refinement. These pseudo-references preserve the real target frames outside the edit mask, while confidence filtering and mixed-data training enable the model to improve without requiring a separate mask-free student or altering its single-pass inference path. Externally generated clips are further incorporated to broaden the visual conditions covered during training.
On HDTF, LiveSync achieves the strongest synchronization scores among the evaluated methods. On the selected AIGC-LipSync subset, it remains competitive in synchronization and achieves strong results on several visual metrics. The ablation studies further confirm the benefit of pseudo-paired supervision. LiveSync also ranks as the second-fastest method in model-side latency under our deployment-oriented protocol, despite using a linear face resolution three times that of the fastest method. Overall, these results suggest that pseudo-reference-based self-training can serve as a complementary route to improving lip-sync models, alongside advances in model architecture.

\clearpage
{
    % \small
    \bibliographystyle{ieeenat_fullname}
    \bibliography{refs,refs/lipsync}
}

\end{document}